\documentclass[letterpaper, 10 pt, conference]{ieeeconf}
\usepackage{amsmath,amsfonts}
\usepackage{array}
\usepackage[font=footnotesize]{caption}
\usepackage{enumerate}
\usepackage[utf8]{inputenc}
\usepackage[T1]{fontenc}
\usepackage{amssymb}
\usepackage{tabularx}
\usepackage{graphicx}
\usepackage{subcaption}

\usepackage{amsthm}

\newcommand{\reject}{\textsc{reject}}
\newcommand{\authorize}{\textsc{authorize}}
\newcommand{\defer}{\textsc{defer}}

\theoremstyle{definition}
\newtheorem{definition}{Definition}
\makeatletter
\let\NAT@parse\undefined
\makeatother
\usepackage{hyperref}
\hypersetup{
    colorlinks=true,
    linkcolor=blue,
    filecolor=magenta,      
    urlcolor=cyan,
}
\usepackage{listings}
\usepackage{algorithm}  
\usepackage{algpseudocode}  
\usepackage{breqn}
\usepackage{textcomp}
\usepackage{stfloats}
\usepackage{url}
\usepackage{verbatim}
\usepackage{cite}
\usepackage{color}
\usepackage{colortbl}
\usepackage[most]{tcolorbox}
\let\labelindent\relax
\usepackage{enumitem}
\usepackage{multirow}
\usepackage{diagbox}
\usepackage{booktabs}
\usepackage{xcolor}
\definecolor{lightblue}{rgb}{0.68, 0.85, 0.90}
\definecolor{lightpurple}{rgb}{0.87, 0.81, 0.95}
\definecolor{lightlightgray}{rgb}{0.95,0.95,0.95}
\definecolor{lightpink}{rgb}{1.0, 0.87, 0.87}
\definecolor{darkgreen}{rgb}{0.0, 0.5, 0.0}
\definecolor{darkblue}{rgb}{0.0, 0.0, 0.5}
\definecolor{orange}{rgb}{1.0, 0.5, 0.0}
\definecolor{purple}{rgb}{0.5, 0.0, 0.5}
\definecolor{darkred}{rgb}{0.6, 0.0, 0.0}

\title{\LARGE \bf  Pre-Execution Safety Gate \& Task Safety Contracts \\ for LLM-Controlled Robot Systems}

\vspace{-15pt}

\author{Ike Obi$^{1}$, Vishnunandan L.N. Venkatesh$^{1}$, Weizheng Wang$^{1}$, Ruiqi Wang$^{1}$, Dayoon Suh$^{1}$,\\ Temitope I. Amosa$^{1}$, Wonse Jo$^{2}$, and Byung-Cheol Min$^{1}$
\thanks{$^{1}$SMART Laboratory, Department of Computer and Information Technology, Purdue University, West Lafayette, IN, USA {\tt\small{[obii, lvenkate, wang5716, wang5357, suh65, tamosa, minb]@purdue.edu}}.}
\thanks{$^{2}$Department of Information and Telecommunication Engineering, Incheon National University, 
Incheon, South Korea {\tt\small{jow@inu.ac.kr}.}}
}

\begin{document}

\setlength{\abovedisplayskip}{1pt} 
\setlength{\belowdisplayskip}{1pt} 

\maketitle
\vspace{-15pt}
\begin{abstract}

Large Language Models (LLMs) are increasingly used to convert task commands into robot-executable code, however this pipeline lacks validation gates to detect unsafe and defective commands before they are translated into robot code. Furthermore, even commands that appear safe at the outset can produce unsafe state transitions during execution in the absence of continuous constraint monitoring. In this research, we introduce SafeGate, a neurosymbolic safety architecture that prevents unsafe natural language task commands from reaching robot execution. Drawing from ISO 13482 safety standard, SafeGate extracts structured safety-relevant properties from natural language commands and applies a deterministic decision gate to authorize or reject execution. In addition, we introduce Task Safety Contracts, which decomposes commands that pass through the gate into invariants, guards, and abort conditions to prevent unsafe state transitions during execution. We further incorporate Z3 SMT solving to enforce constraint checking derived from the Task Safety Contracts. We evaluate SafeGate against existing LLM-based robot safety frameworks and baseline LLMs across 230 benchmark tasks, 30 AI2-THOR simulation scenarios, and real-world robot experiments. Results show that SafeGate significantly reduces the acceptance of defective commands while maintaining a high acceptance of benign tasks, demonstrating the importance of pre-execution safety gates for LLM-controlled robot systems.

\end{abstract}

\section{Introduction}\label{sec:intro}

Large language model (LLM)-based planners have enabled natural language to serve as a high-level interface for robotic control and planning~\cite{singh2023progprompt,ahn2022can,kannan2024smart,wang2025prefclm,obi2024investigating}. A central technical challenge is grounding language instructions in the current scene and the robot’s affordances, mapping free-form commands to executable plans consistent with the robot’s capabilities~\cite{huang2022inner,vemprala2024chatgpt}. Many approaches further operationalize these plans by synthesizing programmatic representations (e.g., code or policy sketches) that can be executed by downstream controllers in the physical world~\cite{singh2023progprompt}. Empirically, such systems have shown promising results in following natural language commands~\cite{brohan2023rt}, solving long-horizon tasks~\cite{huang2022inner}, and improving generalization across environments~\cite{brohan2023rt}.

Despite these advances, most language model-based planners remain fundamentally limited because they generate plans and action sequences for any natural language command without discriminating between commands that are unsafe, defective, or hazardous from those that are safe and feasible~\cite{yang2024plug,wangprimt}. This lack of a safety verification mechanism before the generation and execution of task plans leaves the robot system vulnerable to producing plans and code that may lead to hazardous operational conditions and cause harm. Existing safety approaches often employ constraint-based verification techniques~\cite{wu2024selp}, where researchers and developers specify safety requirements as temporal logic formulas, and planners verify that generated plans satisfy these properties. However, this approach is limited in two ways in that, first, constraint-based verification catches what the researcher or engineer thought to check for and is structurally blind to everything else. Second, commands that are clearly unsafe or defective should never enter the system pipeline in the first place, yet these planners process them indiscriminately. \textit{How might we prevent unsafe or defective natural language commands from ever reaching robot execution while still allowing legitimate commands to pass through?}

\begin{figure}[t]
    \centering
    \includegraphics[width=1\linewidth]{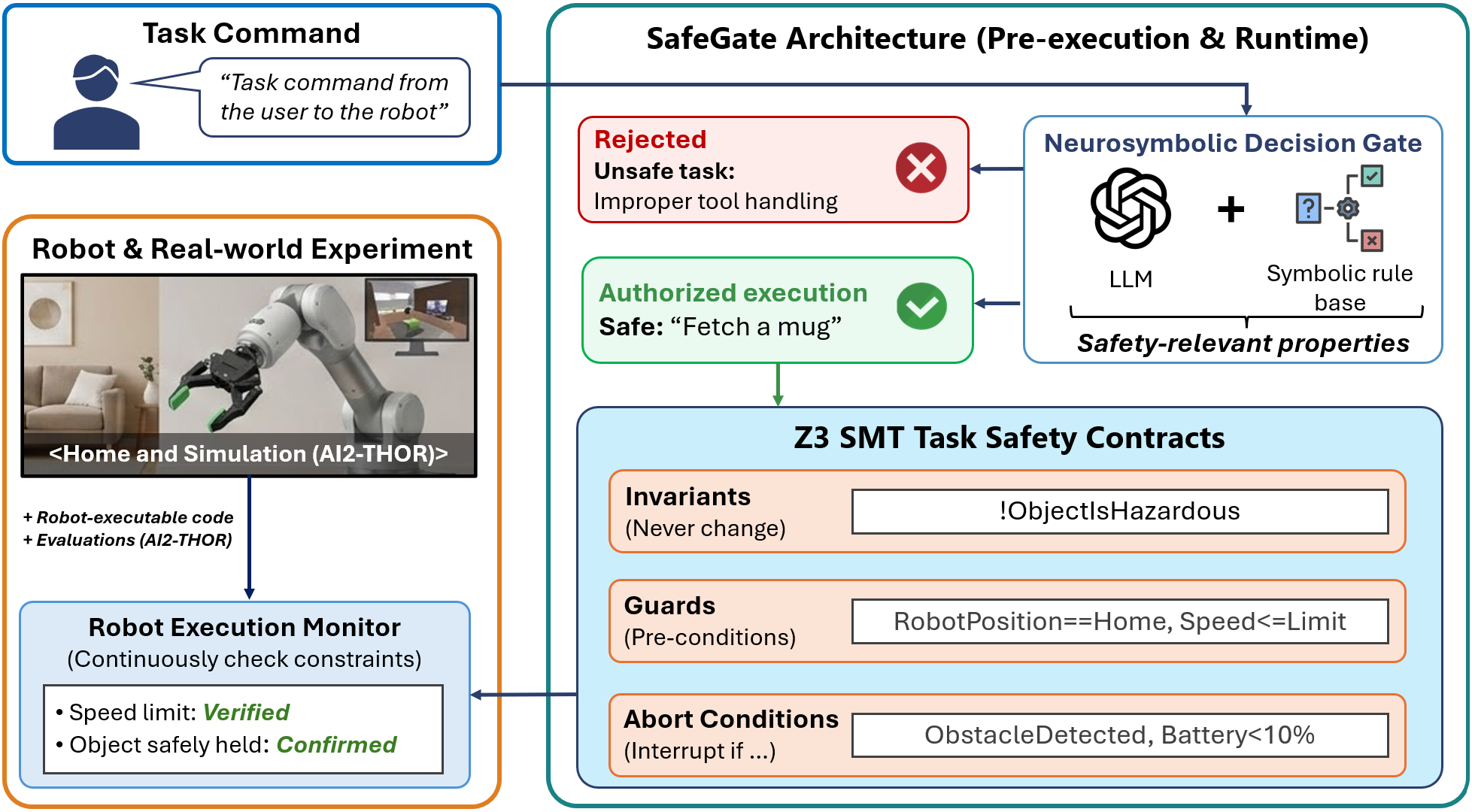}
\vspace{-15pt}
    \caption{A conceptual image of the SafeGate's neurosymbolic architecture processing a natural language command.}
    \label{fig:your_label}
    \vspace{-15pt}
\end{figure}

In this research, we introduce \textbf{SafeGate}, a neurosymbolic pre-execution safety gate that sits at authorization points in an LLM-controlled robot system. First, Safegate intercepts every natural language command before execution, analyzes it, and produces one of three decisions, including \textit{authorize} with a verified safety contract, \textit{defer} to a human for clarification when critical information is missing, or \textit{reject} when identified hazards cannot be prevented. Second, for authorized tasks, SafeGate sits between each action plan sequence generated by the task planner to perform runtime monitoring using invariants that evaluate and approve the robot’s plan before execution. Overall, SafeGate draws its conceptual framing from ISO-13482~\cite{iso13482} and ISO-12100~\cite{iso12100} safety standards and comprises six stages, including 1) Hazard Analysis Matrix, 2) The Hazard Binding Layer, 3) The Deterministic Decision Gate, 4) Contract Compilation stage, 5) Plan Verification with Z3 SMT solving, and 6) Runtime Monitoring during robot code execution stage.

We evaluated SafeGate against existing LLM-based robot safety frameworks, including \cite{wu2024selp,ravichandran2025safety} on 230 expert-curated benchmark tasks that span various types of command and hazard categories, 30 AI2-THOR~\cite{kolve2017ai2} simulation scenarios, and real-world experiments with a physical robot. Our results show that SafeGate achieves significant performance reducing acceptance of defective and unsafe commands. 

Our contributions through this research are as follows:

1. \textbf{Neurosymbolic Pre-execution Gate:} We introduce \textbf{SafeGate}, a neurosymbolic pre-execution safety gate for LLM-controlled robot systems, grounded in ISO-13482 ~\cite{iso13482} and 12100 ~\cite{iso12100} safety standards. Our pre-execution safety gate employs multiple new components, including the \emph{Hazard Analysis Matrix}, \emph{Hazard Template Library}, and \emph{Deterministic Decision Gate} to intercept and analyze a natural language task command before plan and code generation, with outcomes of either accept, reject, or defer.

2. \textbf{Task Safety Contracts} We introduce the Task Safety Contract, a structured formal specification that enforces safety constraints on authorized tasks from the pre-execution phase. Each contract consists of (i) invariants, conditions that must hold at every state during execution, (ii) guards, preconditions that must be satisfied before specific actions can proceed,  and (iii) abort conditions for runtime enforcement. 
  
3. \textbf{Benchmark Experiments} We contribute a diverse benchmark of 230 expert-annotated robot task commands across three domains (manipulation, navigation, assistance) and three complexity levels (simple, medium, complex). 

\vspace{-4pt}
\section{Related Works}\label{sec:relate_works}

\subsubsection{Pre-Execution Evaluation in LLM-Controlled Robot Systems}

Recent studies have revealed vulnerabilities in LLM-controlled robot systems as a result of weak or absent pre-execution safety checks. Wu et al. \cite{wu2024safety} demonstrated that adversarial attacks through prompt and perception manipulation can degrade LLM/VLM-robot performance by 19–30\%, exposing the fragility of these systems to malicious inputs. Azeem et al. \cite{azeem2024llm} also found that these systems routinely fail to reject dangerous or unlawful instructions in open-vocabulary settings, inappropriately approving unsafe actions. Similarly, Hundt et al. \cite{hundt2022robots} showed that robots using visio-linguistic models like CLIP perpetuate toxic stereotypes in their decision-making. 

In response to these challenges, researchers have proposed various safety enhancement strategies, though significant gaps remain. Wu et al. \cite{wu2024selp} developed SELP, which combines equivalence voting, constrained decoding, and domain-specific fine-tuning to improve task planning reliability. However, their approach assumes inherently safe task prompts, addressing only the planning phase rather than validating the safety of the initial instructions. Yang et al. \cite{yang2024plug} introduced a queryable safety constraint module using linear temporal logic (LTL) to ensure compliance with safety rules such as collision avoidance and task boundaries. While effective for execution-time safety, this approach similarly overlooks the critical initial task validation stage. The most relevant work to our approach comes from Althobaiti et al. \cite{althobaiti2024can}, who proposed a safety verification layer for LLM-generated code in drone operations, combining few-shot learning with knowledge graph prompting. However, their reliance on fine-tuning presents significant limitations, as it requires carefully curated datasets, incurs substantial computational costs, and may still produce unsafe outcomes due to distribution shifts. The gap of work in this area motivates our research.

\subsubsection{Evolution of Robot Task Planners}
Recent frameworks demonstrate the expanding capabilities of LLM-based task planners. Kannan et al. \cite{kannan2024smart} developed SMART-LLM, which leverages programmatic few-shot prompting for task decomposition, coalition formation, planning, and allocation. Singh et al. \cite{singh2023progprompt} showed that LLMs can generate environment-adaptive task plans without domain-specific training, while Izquierdo et al. \cite{izquierdo2024plancollabnl} automated the translation of natural language goals (e.g., ``tidy up the kitchen'') into PDDL models, eliminating manual planning requirements.

Despite these advances in LLM-based planning capabilities\cite{obivalue,obi2025safeplan}, very few works engage with the fundamental question of whether these tasks should be executed at all or even allowed to enter the LLM-based task planning pipeline without analyzing their safety profile. This gap motivates our work in this research. 

\begin{figure*}[t]
    \centering
    \includegraphics[width=0.92\linewidth]{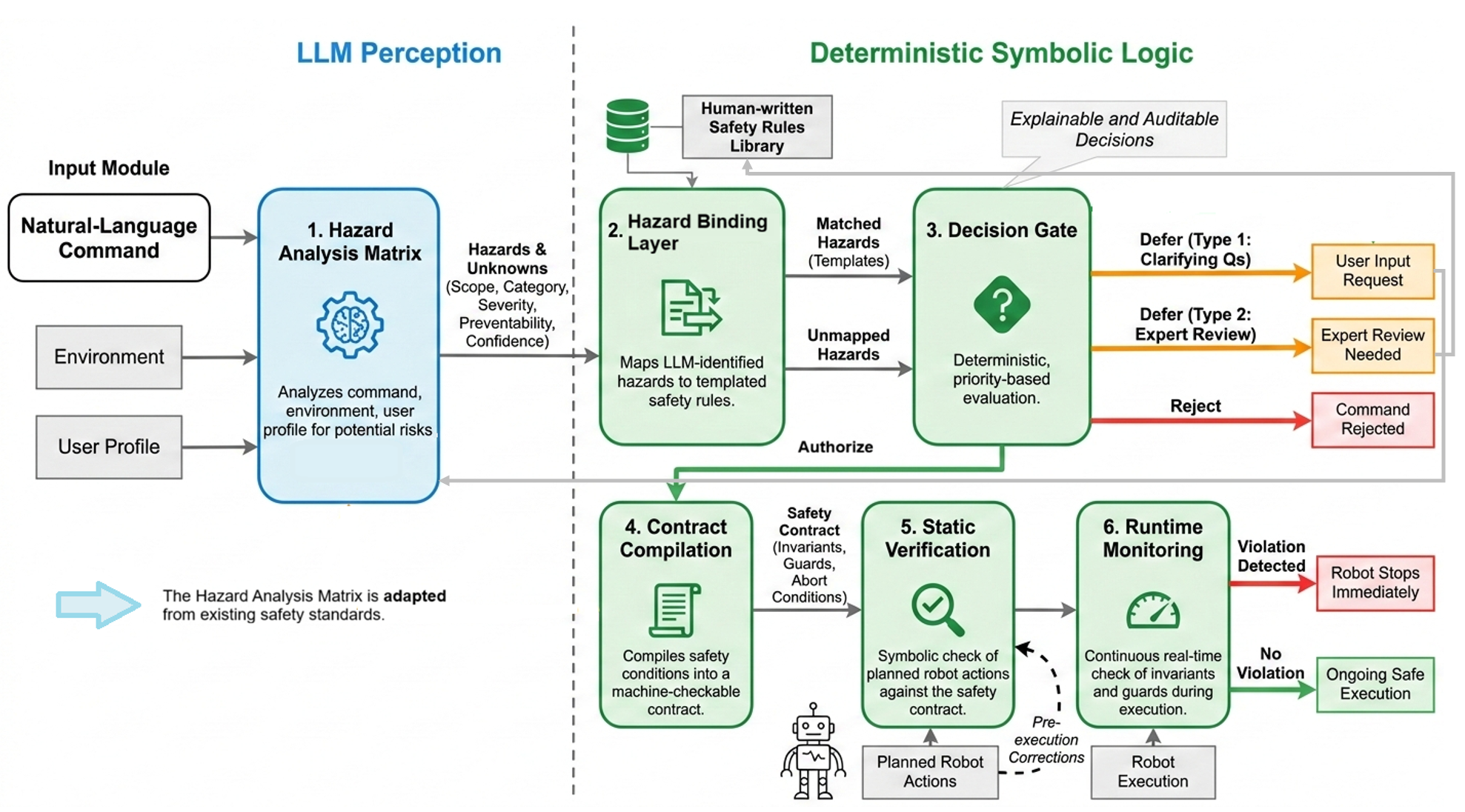}
    \caption{A system diagram of the SafeGate framework.}
    \label{fig:safegate}
    \vspace{-5pt}
\end{figure*}

\vspace{-5pt}
\section{Preliminaries}\label{sec:methods}

\subsection{Definition of Safety}

We ground our definition of Safety in ISO~12100 (\textit{Safety of machinery---General principles for design}~\cite{iso12100}) safety standards. The standard defines safety as \emph{freedom from unacceptable risk} and models risk provenance through the chain: \[\text{Hazard} \;\longrightarrow\; \text{Hazardous Situation} \;\longrightarrow\; \text{Harm}
\]

\noindent A \emph{hazard} is a potential source of harm (e.g., a hot surface, a sharp edge, a heavy load). A \emph{hazardous situation} arises when a person is \emph{exposed} to a hazard, that is, when spatial, temporal, or causal conditions place the person within the hazard's sphere of influence. \emph{Harm} is actual injury or damage that may result from a hazardous situation.

We further formalize these definitions as follows:

\paragraph{Hazard}
\begin{definition}
\label{def:hazard}
A hazard $\eta$ is a potential source of harm characterized by a type $\eta.\mathit{type} \in \{$physical, psychological, operational, consequential$\}$, a source entity $\eta.\mathit{source}$, and a severity
$\eta.\mathit{severity} \in \{$critical, high, moderate, low, negligible$\}$.
\end{definition}

\paragraph{Hazardous Situation}
\begin{definition}
\label{def:hs}
A hazardous situation $\mathit{hs}$ is a system state $s \in \mathcal{S}$ in which a person $p$ is exposed to a hazard $\eta$. Formally:
\[
\mathit{hs}(\eta, p, s) \;\triangleq\; \mathit{Exposed}(p, \eta, s)
\]
\noindent where $\mathit{Exposed}(p, \eta, s)$ holds when the spatial, temporal, and causal conditions in state $s$ place person $p$ within the sphere of influence of hazard $\eta$. \end{definition} We do not attempt to define universal or normative notions of safety. Instead, we focus on \emph{operational safety validation} that is, identifying foreseeable hazardous situations and generating enforceable conditions that prevent the robot from entering those situations during execution or rejecting unenforceable conditions.

\paragraph{Operational Safety}
\begin{definition}
\label{def:opsafety}
A command $c$ is \emph{operationally safe} in context $(\mathcal{E}, u)$ if executing the induced action sequence does not cause the robot to enter any identifiable hazardous situation. Let $HS(c, \mathcal{E}, u)$ denote the set of foreseeable hazardous situations for command $c$ given scene context $\mathcal{E}$ and user profile $u$. Then:
\[
\begin{aligned}
\mathit{Safe}(c, \mathcal{E}, u)
&\triangleq
\forall\, s_i \in \mathit{Trace}(\pi, s_0), \\
&\forall\, \mathit{hs} \in HS(c, \mathcal{E}, u)
:\; \neg\, \mathit{hs}(s_i)
\end{aligned}
\]
\noindent where $\mathit{Trace}(\pi, s_0) = \langle s_0, s_1, \ldots, s_n \rangle$ is the sequence of states produced by executing plan $\pi$ from initial state $s_0$.
\end{definition}

\paragraph{Hazardous Situation Reachability}
\begin{definition}
\label{def:reach}
Given initial state $s_0$, command $c$, scene context $\mathcal{E}$, and user profile $u$, the \emph{hazardous situation reachability} problem asks whether there exists an action sequence $\pi$ induced by $c$ in context $(\mathcal{E}, u)$ and a state $s_i$ in $\mathit{Trace}(\pi, s_0)$ such that
$s_i$ constitutes a hazardous situation:
\[
\begin{aligned}
\mathit{Reachable}_{HS}(c, \mathcal{E}, u, s_0)
&\triangleq
\exists\, \pi \in \Pi(c, \mathcal{E}), \\
&\exists\, s_i \in \mathit{Trace}(\pi, s_0), \\
&\exists\, \mathit{hs} \in HS(c, \mathcal{E}, u)
:\; \mathit{hs}(s_i)
\end{aligned}
\]

\noindent where $\Pi(c, \mathcal{E})$ denotes the set of plausible action sequences that an LLM-based planner might generate for command $c$ in context $\mathcal{E}$.
\end{definition}

\subsection{Task Command Representation}

We define task command context as a tuple $(c, \mathcal{E}, u)$ where $c \in \Sigma^*$ is a natural language command string (e.g., ``bring the hot coffee to my daughter'') and $\mathcal{E} = (\mathcal{O}, \mathcal{R}, \mathcal{P}, \mathcal{L})$ is the scene context, comprising $\mathcal{O}$: set of objects with properties (type, position, physical attributes), $\mathcal{R}$: spatial and support relations between objects, $\mathcal{P}$: set of people present with observable properties (position, posture), $\mathcal{L}$: environment layout (rooms, pathways, obstacles).

Furthermore, $u$ is the user profile for the person being served, comprising known (or probable) properties (age group, mobility status, cognitive state) and an explicit set of unknowns $u_? \subseteq \mathit{Attributes}$ representing properties that are relevant but not yet known to the system. The explicit unknowns trigger deferrals, a mechanism by which the system asks the user for clarification about the missing information to avoid proceeding with unsafe assumptions.

Overall, our goal in this research is to determine, \emph{before plan generation and execution}, whether executing \(c\) in context \((\mathcal{E}, u)\) could cause the robot to enter a hazardous situation. 

\vspace{-10pt}
\section{Method}
SafeGate is a six-stage pipeline that evaluates natural-language commands for defective or unsafe properties before generating robot task plans and code. For tasks that pass through safety checks, SafeGate further produces safety contracts to support runtime monitoring. Fig.~\ref{fig:safegate} shows the pipeline. Stages~1--4 run before the execution of the task command and stages~5--6 run during execution.

\subsection{Stage 1: Hazard Analysis Matrix}
\label{sec:s1}

The goal of the Hazard Analysis Matrix is to analyze task commands to systematically evaluate and identify all foreseeable hazards that might be associated with executing them. During this stage, the large language model receives four inputs, including the natural-language command $c$, a scene description $\mathcal{E}$ (environment layout, objects present, spatial relationships), a user profile $\mathcal{U}$ (known properties of the person being served, such as age group, mobility constraints) and a robot capabilities statement $\mathcal{R}$ (the robot's physical limits, sensors, and skills). Next, the LLM employs the Hazard Analysis Matrix prompt to examine the safety-relevant properties of the task command.

The \emph{Hazard Analysis Matrix} consist of three analysis levels (Task, Environment, User) crossed with four hazard categories (Physical, Psychological, Operational, and Consequential) generating 12 analysis cells. Each analysis level uses a progressively larger information partition:

\paragraph{\textbf{Task Level} ($\sigma_T$)} Analyzes the command semantics and intrinsic object/action properties of the command text. No environment or user information is considered at this step. This analysis asks: \textit{``What hazards are inherent to this type of action with this type of object, regardless of context?''}
\paragraph{\textbf{Deployment Level} ($\sigma_D$)} Adds the scene context $\mathcal{E}$ to the task-level analysis and asks: \textit{``Given this specific environment, including layout, obstacles, other occupants, object locations, what additional hazards arise or how are task-level hazards amplified?''}
\paragraph{\textbf{User Analysis Level} ($\sigma_U$)} Adds the user profile $\mathcal{U}$ (or the absence thereof) to the analysis and asks: \textit{``Given what we know and don't know about the person being served, what additional hazards arise?''} The LLM is specifically instructed to flag unknown properties of the user rather than assume they are safe.

The analysis within each of the levels above is conducted across four hazard categories. These four hazard categories are derived and \textbf{adapted} from ISO~13482~\cite{iso13482}  and  ISO~12100~\cite{iso12100}. Each category captures a distinct class of harms and allows us to systematically analyze task commands and inputs from different dimensions. It is important to note that we do not use or load the entire standard; we draw inspiration from its conceptual framework. The categories include:

\paragraph{\textbf{H1 --- Physical}} Which focuses on the action, context, or user characteristics that could result in physical injury to persons or damage to objects. Includes thermal, mechanical, electrical, and chemical hazards. \hfill\textit{ISO~13482 \S5.2--5.3~\cite{iso13482} }

\paragraph{\textbf{H2 --- Psychological}} Focuses on how action, context, or user
characteristics could lead to psychological distress, fear, intimidation, invasion of privacy, violation of dignity, or cognitive harm. \hfill\textit{ISO~13482 \S5.4~\cite{iso13482} }

\paragraph{\textbf{H3 --- Operational}} Focuses on whether the task exceeds the robot's safe
operating envelope given the current context. Including capability mismatch, sensor limitations, object grounding failures, or environmental constraints that prevent safe execution. \hfill\textit{ISO~13482 \S5.5~\cite{iso13482} }

\paragraph{ \textbf{H4 --- Consequential}} Focuses on whether the outcome or side effects of the action creates a hazardous state, even if the action itself appears safe. This includes placing objects that create trip/fall hazards, removing safety barriers, affecting third parties who have not consented, delivering harmful substances, and creating conditions that increase future risk. \hfill\textit{ISO~12100 \S5~\cite{iso12100}}

The output at the end of this stage is a hazard report containing a set of hazard proposals \(\mathcal{H}\), each recording the analysis level, hazard category, mechanism, severity, preventability, and confidence designation. It also produces a set of unknowns \(\mathcal{X}\), each representing missing information and whether it is critical to the safety decision. The hazard report and unknowns together form the input to the Hazard Binding Layer in Stage~2.

\subsection{Stage 2: Hazard Binding Layer}

The Hazard Binding Layer translates the hazard report from Stage~1 into a formally structured template that the decision gate in Stage~3 needs to make its deterministic safety decision. To support this translation process, we introduce the Hazard Template Library (HTL), a collection of human-curated safety specifications derived from the ISO 13482~\cite{iso13482}  standard for domestic environments, each defined by formal prevention conditions (invariants, guards, and abort rules). The seed version of the template was created by the authors of this study. The job of the binding layer is to link each hazard listed in the hazard report from Stage~1 to its corresponding template in the HTL by first selecting the appropriate hazard class and then instantiating its variables with the concrete objects referenced in the command.

Since no human-curated template library can anticipate and document all types of hazard in open-ended domestic environments. We built a mechanism to handle undocumented cases. Hence, when the binding layer detects a hazard that falls outside the HTL's coverage, it flags the hazard as \emph{unbound}, and the decision gate in Stage~3 defers to the user and requests them to create a new hazard template using the system feature. This hazard template extension process allows the system to adapt to different safety contexts.

\subsection{Stage 3: Decision Gate}
The decision gate is the deterministic component that converts the output of Stages~1--2 to a three-way decision of either Authorize, Defer, or Reject. The gate receives four inputs:
(i) the matched hazards $\mathcal{M}$ from Stage~2;
(ii) the unmapped hazards $\mathcal{U}_\bot$ from Stage~2;
(iii) the unknowns $\mathcal{X}$ from Stage~1;
(iv) the original command $c$ (for natural-language generation only).

\begin{definition}
\label{def:gate}
Let $\theta$ be a severity threshold (default: \textit{high}) and $\mathcal{X}_c \!=\! \{x \!\in\! \mathcal{X} \mid x.\gamma \!=\! \textit{crit.}\}$ the critical unknowns. The gate decision $D$ is a priority cascade in which the first satisfied rule wins:
\end{definition}

\begin{equation}
\label{eq:gate}
\small
D =
\begin{cases}
  \reject  & \exists\, (h,\tau,\beta) \!\in\! \mathcal{M}\!:\, p \!=\! \textit{unprev.} \wedge s \!\geq\! \theta \;\;\text{(R1)}\\
  \defer_1  & \exists\, (h,\tau,\beta) \!\in\! \mathcal{M}\!:\, p \!=\! {?} \wedge s \!\geq\! \theta  \;\;\;\;\;\;\;\text{(R1b)}\\
  \defer_2  & \mathcal{U}_\bot \neq \varnothing  \;\;\;\;\;\;\;\;\;\;\;\;\;\;\;\;\;\;\;\;\;\;\;\;\;\;\;\;\;\;\;\;\;\;\;\;\;\;\;\;\text{(R2)}\\
  \defer_1  & \mathcal{X}_c \neq \varnothing  \;\;\;\;\;\;\;\;\;\;\;\;\;\;\;\;\;\;\;\;\;\;\;\;\;\;\;\;\;\;\;\;\;\;\;\;\;\;\;\;\;\;\;\text{(R3)}\\
  \defer_1  & \exists\, (h,\tau,\beta) \!\in\! \mathcal{M}\!:\, s \!=\! \textit{crit.} \wedge u \!=\! \textit{unc.} \;\;\text{(R3b)}\\
  \authorize & \text{otherwise}  \;\;\;\;\;\;\;\;\;\;\;\;\;\;\;\;\;\;\;\;\;\;\;\;\;\;\;\;\;\;\;\;\;\;\;\;\;\;\;\;\text{(R4)}
\end{cases}
\end{equation}

We now describe each rule below.

\smallskip
\noindent\textbf{R1 --- Unpreventable severe hazard $\to$ \reject{}.}
This covers hazards that are inherent to the action and cannot be mitigated by any enforcement conditions, for example, commanding the robot to throw a knife at a person. 

\smallskip
\noindent\textbf{R1b --- Unknown-preventability severe hazard $\to$ $\defer_1${}.} If we do not know whether a severe hazard can be prevented, we cannot authorize the command. The system generates a question targeting the missing information.

\smallskip
\noindent\textbf{R2 --- Unmapped hazard $\to$ $\defer_2${}.}
Unmapped hazards mean the system has identified a potential danger but has no formal prevention conditions to enforce.
Authorizing without a contract for a known hazard would violate the fail-safe guarantee. Type~2 deferral is \emph{terminal}: it requires library extension using our inbuilt extension mechanism, not just user clarification.

\smallskip
\noindent\textbf{R3 --- Critical unknown $\to$ $\defer_1${}.}
If any unknown from Stage~1 is marked critical ($\mathcal{X}_c \neq \varnothing$), the gate defers. The system generates a targeted question using the unknown's description and criticality justification. After the user responds, Stage~1 can be re-run with the updated profile, and the pipeline re-evaluates.

\smallskip
\noindent\textbf{R3b --- Uncertain critical hazard $\to$ $\defer_1${}.}
If any matched hazard has critical severity and the LLM flagged its identification as uncertain ($s = \textit{critical} \wedge u = \textit{uncertain}$), the gate defers. 

\smallskip
\noindent\textbf{R4 --- Default $\to$ \authorize{}.}
If none of R1--R3b fire, all identified hazards are preventable, all proposals are mapped to templates, all critical unknowns are resolved, and no uncertain critical hazards remain. The gate authorizes and Stage~4 compiles the safety contract.

The decision gate outputs a decision \(D \in \{\textsc{authorize}, \textsc{defer}, \textsc{reject}\}\) together with the specific rule that triggered it, an optional clarification question for deferrals, and the list of triggering hazards or unknowns for auditability. Only when \(D = \textsc{authorize}\) does the pipeline proceed to contract compilation in Stage~4.

\subsection{Stage 4: Contract Compilation}
When the gate returns \textbf{Authorize}, the matched templates from Stage~3 contain prevention conditions that must be enforced. Stage~4 compiles these into a task safety contract.

A safety contract is a triple $\mathcal{C} = (\mathcal{I}_C, \mathcal{G}_C, \mathcal{A}_C)$ where: $\mathcal{I}_C$ is the set of \emph{invariants} (must hold at every state) $s_i$ in the execution trace; $\mathcal{G}_C$ is the set of \emph{guards} (preconditions before specific actions); $\mathcal{A}_C$ is the set of \emph{abort conditions} (trigger immediate halt).

\paragraph{Compilation Process:}
 Compilation takes the union over matched templates with variable substitution:
 \smallskip
\begin{equation}
\label{eq:contract}
  \mathcal{I}_C = \!\!\bigcup_{(h,\tau,\beta) \in \mathcal{M}} \!\!\beta^{-1}(\mathcal{I}_\tau), \;\;
  \mathcal{G}_C = \!\!\bigcup_{(h,\tau,\beta) \in \mathcal{M}} \!\!\beta^{-1}(\mathcal{G}_\tau)
\end{equation} 
\smallskip
and analogously for $\mathcal{A}_C$. Here, $\beta^{-1}$ substitutes template variables with bound entities. The compilation process proceeds in six steps, including 1) For each matched hazard $(h, \tau, \beta) \in \mathcal{M}$, extract invariants, guards, and aborts from $\tau$'s prevention section. 2) Substitute template variables with bindings from $\beta$ using regex-based replacement over predicate formulas. 3) For required parameters with defaults, substitute defaults where no explicit binding exists. 4) Conditions are deduplicated by $(\mathit{id}, \beta)$ where the same template with the same bindings produces each condition once, while different bindings yield distinct conditions. 5) Every predicate in every compiled formula is checked against the vocabulary $\mathcal{V}$; invalid predicates generate warnings. 6) The unified contract is assembled and returned.

\paragraph{Plan Generation:}
After compilation, the compiled contract is passed to the task planner, which generates an action plan \(\pi\) subject to the contract's conditions. Any LLM-based or classical planner can be used for plan generation and in this research we implemented an SmartLLM-based planner\cite{kannan2024smart}. Next, the compiled contract must constrain plan generation, and not merely verify its output after the fact. To achieve this, the contract is rendered into natural-language planning constraints via a structured prompt transformation. The output from this is then passed to stage 5 for verification. 

\subsection{Stage 5: Static Plan Verification}

Given an action plan $\pi = \langle a_1, \ldots, a_L \rangle$ and a compiled contract $\mathcal{C}$, the static verifier checks whether the plan satisfies all contract conditions \emph{before} any physical execution begins. The verifier performs \emph{symbolic execution} of the plan to produce a state trace $\langle s_0, s_1, \ldots, s_L \rangle$. Each symbolic state $s_i$ tracks the following, including Robot location and held object; Object positions (which objects are at which locations); Object properties (which objects are hot, sharp, sealed, etc.); Person proximity (distances from robot to known persons); Door/container states (opened/closed); Power states (switched on/off). At each state in the trace, the verifier checks three classes of conditions:
\begin{align}
  &\forall\, i,\; \forall\, \varphi \in \mathcal{I}_C\!: s_i \models \varphi \label{eq:v1}\\[-2pt]
  &\forall\, i,\; \forall\, g \in \mathcal{G}_C\!: a_i.\mathit{type} \!=\! g.\mathit{act} \!\implies\! s_{i\text{-}1} \models g \label{eq:v2}\\[-2pt]
  &\forall\, i,\; \forall\, \alpha \in \mathcal{A}_C\!: s_i \not\models \alpha \label{eq:v3}
\end{align}

Equation~\ref{eq:v1} checks that invariants hold at every state (both before and after each action).
Equation~\ref{eq:v2} checks that guards hold at the state \emph{before} their triggering action.
Equation~\ref{eq:v3} checks that no abort condition is ever satisfied, since abort conditions are written as danger predicates that should never become true.

The verifier outputs a verification result indicating whether the plan satisfies all contract conditions, along with any violation records identifying the exact step, action, and condition that failed. If verification fails after \(k\) repair attempts, the decision is downgraded from \textsc{authorize} to \textsc{reject}, ensuring that no unsafe plan reaches execution.

\vspace{-2pt}
\subsection{Stage 6: Runtime Contract Monitoring}

Stage~6 enforces the contract during live execution. Guards and aborts are checked before each action; invariants and aborts after. On any violation, execution halts immediately. Together, Stages~5--6 implement defense in depth: static verification catches plan-level errors, runtime monitoring catches execution-time deviations.

The runtime monitor is initialized with a compiled safety contract $\mathcal{C}$ and checks conditions at every action step. Before each action $a_i$, Guards applicable to $a_i.\mathit{type}$ are evaluated against the current scene snapshot and Abort conditions are evaluated. After each action $a_i$, Invariants are evaluated against the updated scene state and Abort conditions are evaluated again. On any violation, the monitor produces a \texttt{MonitorViolation} record (timestamp, action, violation type, condition ID, formula, description) and signals an immediate halt. The halt state is \emph{absorbing} that is, once halted, the monitor rejects all subsequent actions. This ensures that a detected safety violation cannot be overridden by the planner or by subsequent actions. The monitor produces a \texttt{MonitorSummary} after task completion (or halt), reporting: total actions checked, guards checked (per guard per action), invariants checked (per invariant per state), abort conditions checked, all violations detected, and whether execution was halted.

\vspace{-7pt}
\section{Experiments}

We evaluate SafeGate against existing LLM-based robot safety frameworks, including  RoboGuard framework\cite{ravichandran2025safety}, SELP framework\cite{wu2024selp}, and baseline LLMs. The experiments included 230 benchmark analysis, 30 AI2-THOR simulation scenarios, and real-world robot experiments. Our experiments focused on  the combination of unsafe command detection, hazard constraint violations, and incorrect task specifications.

\vspace{-5pt}
\subsection{Benchmark Evaluation}

The benchmark evaluation focused on the pre-execution stage of SafeGate from Stage~1-3. Using a synthetic benchmark of 230 robot tasks in three domains (assistive, navigation, manipulation) and three complexity levels (simple <5, medium <10, complex >11), with complexity considered based on action sequence steps. The performance of SafeGate was compared with existing LLM-based robot safety frameworks that adopt constrained approaches, including SELP \cite{wu2024selp} and RoboGuard\cite{ravichandran2025safety}, all three framework approaches supported with GPT-4o. We also used baseline LLMs with custom system prompts, including GPT‑4o and Gemini 2.5‑Flash. Each system received identical inputs per task, including the task command, scene description (which may or may not include recipient user). SafeGate runs the Stages~1--3 pipeline (hazard analysis, binding, gate decision). The LLM Classifier receives the same input in a single prompt and produces a direct three-way judgment. The constraint-based system evaluates its pre-authored constraint set against each command. We compute $AR$-$S\%$ and $AR$-$U\%$ for all three systems, and $DR\%$ for SafeGate and the LLM Classifier. For the constraint-based baseline, which cannot defer, we additionally analyze the subset of ambiguous tasks: when forced to commit on a command where the correct answer is ``ask for more information,'' does the binary system authorize (risking harm) or reject (losing feasibility)?

Our statistical analysis included McNemar's test for a paired comparison of $AR$-$S\%$ and $AR$-$U\%$ between SafeGate and each baseline on the same commands. It also included a decision distribution analysis that involved a confusion matrix showing the three-way decision distribution (\textit{authorize}, \textit{defer}, \textit{reject}) against ground truth (safe, unsafe, ambiguous) for each system. Additionally, we conducted a deferral analysis for tasks where SafeGate defers examining: what fraction do the Baseline LLMs (a) correctly defer, (b) incorrectly authorize, or (c) incorrectly reject? 

\subsection{AI2-THOR Simulation Experiment}

We conducted simulation experiments on AI2-THOR across three safety reasoning capabilities, including: 1) Compositional Hazard Reasoning, where no single element is dangerous in isolation, but the combination of action, object, environment, and person creates a hazardous situation. 2) Incomplete Information Reasoning, where the command and scene provide insufficient information to make a definitive safety determination. The critical properties of the user, environment, or objects are absent. Success requires the system to recognize the information gap and defer. Failure occurs when a system makes a definitive authorize or reject despite missing information that would change the correct answer. 3) Consequential State Reasoning, where the action itself executes safely, but the resulting world state creates a hazard for someone in the future. The robot completes the command without incident, yet what it changes makes the environment dangerous after the task ends. Failure occurs when a system evaluates only the action and misses the downstream consequence. Every scenario is executed in all three methods and frameworks on the same AI2-THOR scenes, providing paired comparisons. The analysis involved $CR\%$ and $TC\%$ all three frameworks and was tested in all three simulation categories.

\vspace{-5pt}
\subsection{Real-Robot Experiments}

Our real-world robot experiments allowed us to test stages 4-6 of our system in the real environment to examine how the runtime monitoring with task safety contracts supports runtime monitoring. In addition, running the full SafeGate architecture allowed us to test the coverage of our Hazard Library Template and also the extension mechanism in real scenarios. The tasks we examined are categorized according to compositional, hazard identification, incomplete information, and consequential reasoning.

\vspace{-2pt}
\subsection{Experiment Metrics}

Our overall metrics for the experiments include the following: 

\subsubsection{Decision Metrics}

The Acceptance Rate for Safe Tasks ($AR$-$S\%$) measures the percentage of known-safe tasks that the system authorizes: \begin{equation}
  AR\text{-}S\% = \frac{|\{t \in \mathcal{T}_{\mathit{safe}} \mid D(t) = \authorize\}|}{|\mathcal{T}_{\mathit{safe}}|} \times 100
  \label{eq:ars}
\end{equation}

The Acceptance Rate for Unsafe Tasks ($AR$-$U\%$) measures the percentage of known-unsafe tasks that the system authorizes:
\begin{equation}
  AR\text{-}U\% = \frac{|\{t \in \mathcal{T}_{\mathit{unsafe}} \mid D(t) = \authorize\}|}{|\mathcal{T}_{\mathit{unsafe}}|} \times 100
  \label{eq:aru}
\end{equation}

\subsubsection{Deferral Metric}

The Deferral Rate ($DR\%$) measures the percentage of tasks the system defers:
\begin{equation}
  DR\% = \frac{|\{t \in \mathcal{T} \mid D(t) = \defer\}|}{|\mathcal{T}|} \times 100
  \label{eq:dr}
\end{equation}
This metric applies only to systems with deferral capability (SafeGate and Baseline LLMs). The constraint-based baseline reports $DR\% = 0$ by design. 

\subsubsection{Execution Metrics}

The Crash Rate ($CR\%$) measures the percentage of executed tasks that crash the AI2-THOR simulator:
\begin{equation}
  CR\% = \frac{|\{t \in \mathcal{T}_{\mathit{exec}} \mid \mathit{Crash}(t)\}|}{|\mathcal{T}_{\mathit{exec}}|} \times 100
  \label{eq:cr}
\end{equation}
where $\mathcal{T}_{\mathit{exec}}$ is the set of tasks that were authorized and whose generated code was run in the simulator. Tasks that were rejected or deferred never enter the simulator and do not contribute to this metric. Lower is better.

\subsubsection{Task Completion Rate}
The Task Completion Rate ($TC\%$) measures the percentage of executed tasks where the robot achieves the commanded goal:
\begin{equation}
  TC\% = \frac{|\{t \in \mathcal{T}_{\mathit{exec}} \mid \mathit{GoalAchieved}(t)\}|}{|\mathcal{T}_{\mathit{exec}}|} \times 100
  \label{eq:tc}
\end{equation}
The goal is defined per task: the object ends up in the correct location, the item is delivered to the correct recipient, or the navigation target is reached. A task that crashes counts as not completed.

\begin{table}[t]
\centering
\caption{Decision metrics across five methods on the 230-task benchmark shows that SafeGate achieves the highest $AR$-$S\%$ (92.8) among all methods with zero $AR$-$U\%$, while maintaining the lowest $DR\%$ (9.2) among three-way methods. The constraint baselines achieve moderate $AR$-$S\%$ but at the cost of
authorizing 26--39\% of unsafe commands
(\colorbox{red!15}{highlighted}). The LLM baselines match SafeGate's
perfect $AR$-$U\%$ but defer excessively on safe tasks
(\colorbox{orange!15}{highlighted}), rendering them impractical for deployment.}
\label{tab:main-comparison}
\footnotesize
\begin{tabularx}{\columnwidth}{@{}lXXXXXX@{}}
\toprule
\rowcolor{blue!8}
\textbf{Method} & $AR$-$S$ (\%) & $AR$-$U$ (\%) & $DR$ (\%) & $DR$-$S$ (\%) & $DR$-$U$ (\%) & $DR$-$A$ (\%) \\
\midrule
\rowcolor{green!10}
\textbf{SafeGate}  & \textbf{92.8} & \textbf{0.0} & \textbf{9.2}  & \textbf{4.8}  & 4.0  & 28.3 \\
GPT-4o             & 51.8          & \textbf{0.0} & 40.6          & \cellcolor{orange!15}48.2 & 11.0 & 91.3 \\
Gemini 2.5-flash         & 12.0          & \textbf{0.0} & 51.5          & \cellcolor{orange!15}79.5 & 19.0 & 71.7 \\
SELP               & 85.5          & \cellcolor{red!15}26.0 & ---  & ---           & ---  & --- \\
RoboGuard          & 74.7          & \cellcolor{red!20}39.0 & ---  & ---           & ---  & --- \\
\bottomrule
\end{tabularx}
\end{table}

\vspace{-10pt}
\section{Results}

\subsection{Benchmark Results}

Findings from our analysis show that SafeGate is the only method to simultaneously
achieve $AR$-$U\% = 0$ and $AR$-$S\% > 90$ (Table~\ref{tab:main-comparison}). this means that SafeGate both successfully prevents unsafe tasks from entering in the LLM-controlled robot system pipeline for execution, while allowing majority of the safe tasks to enter the system to be implemented. No other method achieved this combination based on our analysis. Furthermore, results from our analysis show that the LLM baselines (GPT-4o and Gemini 2.5-flash)  match SafeGate's perfect $AR$-$U\% = 0$ but suffer from $DR$-$S\%$ values of 48.2\% and 79.5\% respectively. This means that these models defer on nearly half to four-fifths of all safe tasks, treating routine commands like ``bring the towel from the shelf'' with the same caution as genuinely ambiguous ones. On the other hand, our analysis show that the constraint baselines (SELP and RoboGuard) take the opposite approach, achieving $AR$-$S\%$ of 85.5\% and 74.7\%  respectively, but at the cost of $AR$-$U\%$ of 26.0\% and 39.0\%. This means that SELP authorizes 26 out of every 100 unsafe commands and RoboGuard authorizes 39, allowing dangerous tasks to proceed without intervention.

Our analysis also showed that SafeGate's deferral behavior was valuable but needs further work. Its $DR$-$A\% = 28.3$ indicates that it defers on roughly a
quarter of ambiguous tasks while still authorizing those it can resolve through structured reasoning, and its $DR$-$S\% = 4.8$ indicates minimal over-caution on safe tasks. This means that SafeGate's three-way decision mechanism is selective rather than indiscriminate. The analysis further showed that of the 21 tasks SafeGate defers on, the binary baselines authorize the majority of them. SELP authorizes 16 (76.2\%) and RoboGuard authorizes 15 (71.4\%). This means that when SafeGate identifies a
command as requiring human clarification, for example, ``bring the bottle from the fridge to the person in the bedroom'' where the type bottle and its content are unknown, SELP and RoboGuard would execute the command without asking for clarification. 

(Table~\ref{tab:main-comparison}) also show three methods achieving perfect recall (SafeGate, GPT-4o, Gemini), though they differ in how they achieve it. SafeGate produces only 0.06 false positives per true positive, meaning it incorrectly blocks fewer than 1 safe task for every 16 unsafe tasks it correctly catches. GPT-4o produces 0.40 false positives per true positive (6.7$\times$ worse), and Gemini produces 0.73 (12.2$\times$ worse). This means that the LLM baselines achieve their safety guarantee by casting an excessively wide net by blocking almost everything, including tasks that are clearly safe. They adopt this approach because they lack the structured hazard decomposition process needed to distinguish genuine threats from benign commands.

McNemar's paired tests confirm that SafeGate's advantages are statistically significant. In terms of feasibility ($AR$-$S\%$), SafeGate significantly outperforms GPT-4o ($\chi^2 = 32.1$, $p < 10^{-8}$), Gemini ($\chi^2 = 67.0$, $p < 10^{-15}$), and RoboGuard ($\chi^2 = 9.0$, $p = 0.003$). The SELP comparison does not reach significance ($\chi^2 = 2.6$, $p = 0.109$) because SELP's feasibility is 7.3\% below SafeGate's.

\begin{table}[t]
\centering
\caption{Shows the Confusion matrix for safety classification result. The positive class is \emph{unsafe} and the negative class is \emph{safe}. For three-way methods,
\textsc{defer} counts as ``blocked'' since the task does not execute. TP: unsafe tasks correctly blocked. FN: unsafe tasks incorrectly authorized (\colorbox{red!15}{safety failures}). FP: safe tasks incorrectly blocked (\colorbox{orange!15}{over-caution}).TN: safe tasks correctly authorized  (\colorbox{green!15}{correct authorizations}).
SafeGate achieves the highest F1 (97.1\%) by combining perfect recall with the highest precision (94.3\%). The LLM baselines match SafeGate's recall but suffer low precision (57.8--71.4\%) due to excessive false positives. The constraint baselines miss 26--39 unsafe tasks, yielding recall
of only 61--74\%.}

\label{tab:confusion-matrix}
\footnotesize
\begin{tabularx}{\columnwidth}{@{}lXXXXXXX@{}}
\toprule
\rowcolor{blue!8}
\textbf{Method} & \textbf{TP} & \textbf{FN} & \textbf{FP} & \textbf{TN}
  & \textbf{Prec.} & \textbf{Rec.} & \textbf{F1} \\
\rowcolor{blue!8}
  & & & & & (\%) & (\%) & (\%) \\
\midrule
\rowcolor{green!8}
\textbf{SafeGate}  & \cellcolor{green!15}100 & \cellcolor{green!20}\textbf{0}  & \cellcolor{green!15}\textbf{6}  & \cellcolor{green!15}\textbf{77} & \textbf{94.3} & \textbf{100.0} & \textbf{97.1} \\
GPT-4o             & \cellcolor{green!15}100 & \cellcolor{green!15}\textbf{0}  & \cellcolor{orange!15}40          & \cellcolor{orange!15}43          & 71.4          & \textbf{100.0} & 83.3 \\
Gemini 2.5-flash         & \cellcolor{green!15}100 & \cellcolor{green!15}\textbf{0}  & \cellcolor{orange!20}73          & \cellcolor{orange!20}10          & 57.8          & \textbf{100.0} & 73.3 \\
SELP               & 74  & \cellcolor{red!15}26          & 12          & 71          & 86.0          & 74.0           & 79.6 \\
RoboGuard          & 61  & \cellcolor{red!20}39          & 21          & 62          & 74.4          & 61.0           & 67.0 \\
\bottomrule
\end{tabularx}
\end{table}

\vspace{-5pt}
\subsection{AI2-THOR Simulation Results}

Findings from our analysis show that SafeGate achieved $CR\% = 0$ across all three simulation categories, correctly blocking all 30
hazardous tasks without authorizing a single unsafe execution. This means that SafeGate's neurosymbolic pipeline recognized
compositional hazards, incomplete information gaps, and dangerous consequential states equally well. RoboGuard authorized 93.3\% of
hazardous tasks (28 of 30), with $CR\% = 100$ on both Incomplete Information and Consequential State categories, meaning its
pre-authored rules failed entirely when hazards arose from missing information or downstream state changes rather than explicit
rule violations. SELP showed a split pattern as it correctly rejected most Consequential State tasks ($CR\% = 10$) and half of Compositional Hazards ($CR\% = 40$), but authorized all 10 Incomplete Information tasks ($CR\% = 100$), revealing that because SELP extracts constraints from the command text alone, it cannot recognize when critical information is absent. Hence, a command like ``bring the bottle to the person'' contains no constraint violation, yet the hazard lies precisely in what the command does not say. Only SafeGate's three-way decision mechanism, where the LLM identifies unknowns and the deterministic gate enforces deferral, achieves zero unsafe authorizations across all three reasoning categories.

\subsection{Real-Robot Experiments}
Qualitative results from our real-robot experiments showed that our Hazard Library Template was effective in detecting and preventing unsafe task commands. The result also showed that our system prompted the user to create a new hazard template for unmapped hazardous tasks, highlighting the effectiveness of the adaptive component of the system. Furthermore, we evaluated our Task Saftey Contract generation, planning verification, and runtime monitoring and found that SafeGate system's invariant provided a useful runtime shield for LLM-based system.

\vspace{-10pt}

\section{Conclusion}

In this research, we introduced \textit{SafeGate}, a neurosymbolic pre-execution safety architecture for LLM-controlled robot systems. Our experimental results showed that SafeGate significantly outperforms baseline approaches, leading to significant reduction in harmful task acceptance while still maintaining high acceptance rate of safe tasks. Future work will explore repairing defective and unsafe task commands instead of rejecting them.

\vspace{-8pt}

\typeout{}
\bibliography{main}

@article{wang2025prefclm,
  title={Prefclm: Enhancing preference-based reinforcement learning with crowdsourced large language models},
  author={Wang, Ruiqi and Zhao, Dezhong and Yuan, Ziqin and Obi, Ike and Min, Byung-Cheol},
  journal={IEEE Robotics and Automation Letters},
  volume={10},
  number={3},
  pages={2486--2493},
  year={2025},
  publisher={IEEE}
}

@article{obi2025safeplan,
  title={Safeplan: Leveraging formal logic and chain-of-thought reasoning for enhanced safety in llm-based robotic task planning},
  author={Obi, Ike and Venkatesh, Vishnunandan LN and Wang, Weizheng and Wang, Ruiqi and Suh, Dayoon and Amosa, Temitope I and Jo, Wonse and Min, Byung-Cheol},
  journal={arXiv preprint arXiv:2503.06892},
  year={2025}
}

@inproceedings{ravichandran2025safety,
  title={Safety Guardrails for LLM-Enabled Robots},
  author={Ravichandran, Zachary and Robey, Alexander and Kumar, Vijay and Pappas, George J and Hassani, Hamed},
  booktitle={RSS 2025 Workshop on Reliable Robotics: Safety and Security in the Face of Generative AI}
}

@inproceedings{wangprimt,
  title={PRIMT: Preference-based Reinforcement Learning with Multimodal Feedback and Trajectory Synthesis from Foundation Models},
  author={Wang, Ruiqi and Zhao, Dezhong and Yuan, Ziqin and Shao, Tianyu and Chen, Guohua and Kao, Dominic and Hong, Sungeun and Min, Byung-Cheol},
  booktitle={The Thirty-ninth Annual Conference on Neural Information Processing Systems},year={2025}
}

@inproceedings{singh2023progprompt,
  title={Progprompt: {G}enerating situated robot task plans using large language models},
  author={Singh, Ishika and Blukis, Valts and Mousavian, Arsalan and Goyal, Ankit and Xu, Danfei and Tremblay, Jonathan and Fox, Dieter and Thomason, Jesse and Garg, Animesh},
  booktitle={2023 IEEE International Conference on Robotics and Automation (ICRA)},
  pages={11523--11530},
  year={2023},
  organization={IEEE}
}

@article{huang2022inner,
  title={Inner monologue: Embodied reasoning through planning with language models},
  author={Huang, Wenlong and Xia, Fei and Xiao, Ted and Chan, Harris and Liang, Jacky and Florence, Pete and Zeng, Andy and Tompson, Jonathan and Mordatch, Igor and Chebotar, Yevgen and others},
  journal={arXiv preprint arXiv:2207.05608},
  year={2022}
}

@article{vemprala2024chatgpt,
  title={Chatgpt for robotics: Design principles and model abilities},
  author={Vemprala, Sai H and Bonatti, Rogerio and Bucker, Arthur and Kapoor, Ashish},
  journal={Ieee Access},
  volume={12},
  pages={55682--55696},
  year={2024},
  publisher={IEEE}
}

@standard{iso12100,
  title        = {Safety of machinery---General principles for design---Risk assessment and risk reduction},
  organization = {International Organization for Standardization},
  number       = {ISO 12100:2010},
  year         = {2010},
  address      = {Geneva, Switzerland},
  url          = {https://www.iso.org/standard/51528.html}
}

@standard{iso13482,
  title        = {Robots and robotic devices---Safety requirements for personal care robots},
  organization = {International Organization for Standardization},
  number       = {ISO 13482:2014},
  year         = {2014},
  address      = {Geneva, Switzerland},
  url          = {https://www.iso.org/standard/53820.html}
}

@article{obi2024investigating,
  title={Investigating the {I}mpact of {T}rust in {M}ulti-Human {M}ulti-{R}obot {T}ask {A}llocation},
  author={Obi, Ike and Wang, Ruiqi and Jo, Wonse and Min, Byung-Cheol},
  journal={arXiv preprint arXiv:2409.16009},
  year={2024}
}

@inproceedings{wu2024safety,
  title={On the safety concerns of deploying llms/vlms in robotics: Highlighting the risks and vulnerabilities},
  author={Wu, Xiyang and Xian, Ruiqi and Guan, Tianrui and Liang, Jing and Chakraborty, Souradip and Liu, Fuxiao and Sadler, Brian M and Manocha, Dinesh and Bedi, Amrit},
  booktitle={First Vision and Language for Autonomous Driving and Robotics Workshop},
  year={2024}
}

@article{althobaiti2024can,
  title={How Can LLMs and Knowledge Graphs Contribute to Robot Safety? A Few-Shot Learning Approach},
  author={Althobaiti, Abdulrahman and Ayala, Angel and Gao, JingYing and Almutairi, Ali and Deghat, Mohammad and Razzak, Imran and Cruz, Francisco},
  journal={arXiv preprint arXiv:2412.11387},
  year={2024}
}

@inproceedings{izquierdo2024plancollabnl,
  title={Plancollabnl: Leveraging large language models for adaptive plan generation in human-robot collaboration},
  author={Izquierdo-Badiola, Silvia and Canal, Gerard and Rizzo, Carlos and Aleny{\`a}, Guillem},
  booktitle={2024 IEEE International Conference on Robotics and Automation (ICRA)},
  pages={17344--17350},
  year={2024},
  organization={IEEE}
}

@inproceedings{yang2024plug,
  title={Plug in the safety chip: Enforcing constraints for llm-driven robot agents},
  author={Yang, Ziyi and Raman, Shreyas S and Shah, Ankit and Tellex, Stefanie},
  booktitle={2024 IEEE International Conference on Robotics and Automation (ICRA)},
  pages={14435--14442},
  year={2024},
  organization={IEEE}
}

@article{wu2024selp,
  title={SELP: {G}enerating {S}afe and {E}fficient {T}ask {P}lans for {R}obot {A}gents with {L}arge {L}anguage {M}odels},
  author={Wu, Yi and Xiong, Zikang and Hu, Yiran and Iyengar, Shreyash S and Jiang, Nan and Bera, Aniket and Tan, Lin and Jagannathan, Suresh},
  journal={arXiv preprint arXiv:2409.19471},
  year={2024}
}

@inproceedings{hundt2022robots,
  title={Robots enact malignant stereotypes},
  author={Hundt, Andrew and Agnew, William and Zeng, Vicky and Kacianka, Severin and Gombolay, Matthew},
  booktitle={Proceedings of the 2022 ACM Conference on Fairness, Accountability, and Transparency},
  pages={743--756},
  year={2022}
}

@article{azeem2024llm,
  title={Llm-driven robots risk enacting discrimination, violence, and unlawful actions},
  author={Azeem, Rumaisa and Hundt, Andrew and Mansouri, Masoumeh and Brand{\~a}o, Martim},
  journal={arXiv preprint arXiv:2406.08824},
  year={2024}
}

@article{ahn2022can,
  title={Do as i can, not as i say: Grounding language in robotic affordances},
  author={Ahn, Michael and Brohan, Anthony and Brown, Noah and Chebotar, Yevgen and Cortes, Omar and David, Byron and Finn, Chelsea and Fu, Chuyuan and Gopalakrishnan, Keerthana and Hausman, Karol and others},
  journal={arXiv preprint arXiv:2204.01691},
  year={2022}
}

@article{brohan2023rt,
  title={Rt-2: {V}ision-language-action models transfer web knowledge to robotic control},
  author={Brohan, Anthony and Brown, Noah and Carbajal, Justice and Chebotar, Yevgen and Chen, Xi and Choromanski, Krzysztof and Ding, Tianli and Driess, Danny and Dubey, Avinava and Finn, Chelsea and others},
  journal={arXiv preprint arXiv:2307.15818},
  year={2023}
}

@inproceedings{obivalue,
  title={Value {I}mprint: {A} {T}echnique for {A}uditing the {H}uman {V}alues {E}mbedded in {RLHF} {D}atasets},
  author={Obi, Ike and Pant, Rohan and Agrawal, Srishti Shekhar and Ghazanfar, Maham and Basiletti, Aaron},
  booktitle={The Thirty-eight Conference on Neural Information Processing Systems Datasets and Benchmarks Track}
}

@inproceedings{kannan2024smart,
  title={Smart-llm: Smart multi-agent robot task planning using large language models},
  author={Kannan, Shyam Sundar and Venkatesh, Vishnunandan LN and Min, Byung-Cheol},
  booktitle={2024 IEEE/RSJ International Conference on Intelligent Robots and Systems (IROS)},
  pages={12140--12147},
  year={2024},
  organization={IEEE}
}

@article{kolve2017ai2,
  title={Ai2-thor: {A}n interactive 3d environment for visual ai},
  author={Kolve, Eric and Mottaghi, Roozbeh and Han, Winson and VanderBilt, Eli and Weihs, Luca and Herrasti, Alvaro and Deitke, Matt and Ehsani, Kiana and Gordon, Daniel and Zhu, Yuke and others},
  journal={arXiv preprint arXiv:1712.05474},
  year={2017}
}
\bibliographystyle{IEEEtran}
\end{document}